\documentclass{article}

\PassOptionsToPackage{numbers,compress}{natbib}

\usepackage[preprint]{neurips_ai4d3_2025}

\usepackage[utf8]{inputenc} 
\usepackage[T1]{fontenc}    
\usepackage{hyperref}       
\usepackage{url}            
\usepackage{booktabs}       
\usepackage{amsfonts}       
\usepackage{nicefrac}       
\usepackage{microtype}      
\usepackage{xcolor}         

\usepackage{amsmath}
\usepackage{amssymb}
\usepackage{mathtools}
\usepackage{amsthm}

\title{Beyond Atoms: Evaluating Electron Density Representation for 3D Molecular Learning}

\author{%
  Patricia Suriana \\
  Prescient Design, Genentech\\
  SSF, USA \\
  \texttt{surianap@gene.com} \\
  \And
  Joshua A. Rackers \\
  Prescient Design, Genentech\\
  SSF, USA \\
  \And
  Ewa M. Nowara \\
  Prescient Design, Genentech\\
  SSF, USA \\
  \And
  Pedro O. Pinheiro \\
  Prescient Design, Genentech\\
  SSF, USA \\
  \And
  John M. Nicoloudis \\
  Structural Biology, Genentech \\
  SSF, USA
  \And
  Vishnu Sresht \\
  Prescient Design, Genentech\\
  SSF, USA \\
}

\begin{document}

\maketitle

\begin{abstract}
Machine learning models for 3D molecular property prediction typically rely on atom-based representations, which may overlook subtle physical information. Electron density maps—the direct output of X-ray crystallography and cryo-electron microscopy—offer a continuous, physically grounded alternative. We compare three voxel-based input types for 3D convolutional neural networks (CNNs): atom types, raw electron density, and density gradient magnitude, across two molecular tasks—protein–ligand binding affinity prediction (PDBbind) and quantum property prediction (QM9). We focus on voxel-based CNNs because electron density is inherently volumetric, and voxel grids provide the most natural representation for both experimental and computed densities. On PDBbind, all representations perform similarly with full data, but in low-data regimes, density-based inputs outperform atom types, while a shape-based baseline performs comparably—suggesting that spatial occupancy dominates this task. On QM9, where labels are derived from Density Functional Theory (DFT) but input densities from a lower-level method (XTB), density-based inputs still outperform atom-based ones at scale, reflecting the rich structural and electronic information encoded in density. Overall, these results highlight the task- and regime-dependent strengths of density-derived inputs, improving data efficiency in affinity prediction and accuracy in quantum property modeling.
\end{abstract}

\section{Introduction}
\label{introduction}

Machine learning (ML) has become an essential component of structure-based small molecule discovery, supporting tasks such as virtual screening, lead optimization, and protein–ligand binding affinity prediction \citep{Jimenez2018KDEEP, Ragoza2017}. A critical challenge in these applications is selecting a representation that effectively captures the physical and chemical complexity of molecular systems. Most current methods rely on atom-based features, using atomic coordinates and element types extracted from experimentally resolved or computationally modeled structures.

Although widely used, atom-based representations abstract away important physical details. Atomic coordinates are not measured directly; they are inferred by fitting an atomistic model to experimental electron density maps obtained from X-ray crystallography or cryo-electron microscopy. This model-building process depends on expert interpretation and heuristic refinement procedures, introducing potential bias and error. Structural inaccuracies in public databases such as the Protein Data Bank (PDB) \citep{berman2000protein} are well documented \citep{dauter2014avoidable, wlodawer2018detect}. Moreover, by discretizing molecules into point-like atoms, these representations ignore the continuous distribution of electron density that governs molecular interactions.

Electron density maps, the direct output of X-ray crystallography and cryo-electron microscopy (cryo-EM), offer a more direct and physically grounded alternative. These maps represent a 3D scalar field describing the spatial distribution of electrons. Unlike fitted atomistic models, they encode both the extent and overlap of electron clouds and inherently capture features such as conformational heterogeneity and structural uncertainty, which manifest as diffuse or mixed density in flexible regions.

Electron density has several properties that make it appealing for ML applications:
\begin{itemize}
\item \textbf{Directly derived from experiment:} Density maps are obtained directly from physical measurements, bypassing the lossy step of atomic model fitting \citep{phenix}.
\item \textbf{Continuous encoding of interactions:} Because molecular forces arise from electron distribution, density may enable more physically faithful modeling of interaction strength and geometry.
\item \textbf{Implicit representation of flexibility and uncertainty:} Conformational variability appears naturally in the map, without additional modeling assumptions.
\end{itemize}

These characteristics suggest that density-based representations may offer a richer signal for learning molecular properties—particularly those governed by electronic structure and interactions, such as binding affinity and quantum mechanical properties. Intuitively, one might expect this richer information to translate into improved model performance, especially under data-limited conditions.

At the same time, atom-type representations provide strong chemical priors by explicitly labeling atoms with their identities (e.g., C, N, O). These priors embed known chemical patterns and may be especially useful in high-data or chemically diverse regimes. This raises a central question: when do density-based representations offer an advantage over atom-based ones?

To address this, we compare three voxel-based representations for 3D convolutional neural networks (CNNs): atom-type channels, raw electron density, and the gradient magnitude of density. The gradient captures rapid spatial changes in density and may highlight features relevant to chemical interactions, such as bonding regions or non-covalent contacts \citep{Wang2022DensityGen}. We use 3D CNNs because they are well suited for volumetric data, as electron density forms a continuous field in three-dimensional space. Although computed densities (e.g., from XTB or DFT) can be expressed in alternative bases, voxelization provides a unified spatial framework for both experimental and theoretical sources. Because our objective is to benchmark different voxel representations rather than optimize architectures, we focus on CNNs, which can directly process volumetric data, whereas graph- or transformer-based networks \citep{Gilmer2017, Schutt2017NIPS, Thomas2018, Fuchs2020} cannot natively represent continuous 3D density fields. This makes 3D CNNs a natural architectural choice for evaluating volumetric representations.

We evaluate these representations on two tasks: (1) protein–ligand binding affinity prediction using the PDBbind dataset, and (2) quantum property prediction for small molecules using QM9. These tasks differ in physical scale, label origin, and modeling assumptions, providing complementary perspectives. Specifically, we ask: Do density-based inputs improve model performance in low-data settings? Do these benefits persist at scale, even when the input densities are approximate or noisy? By analyzing how representation interacts with data regime and model capacity, we aim to clarify the conditions under which density-derived inputs improve 3D molecular learning.

\section{Related Work}
\label{sec:related_work}

\paragraph{Structure-Based Learning with Atomic Coordinates.}
Most machine learning models for structure-based molecular discovery rely on atomic coordinate–based representations. Graph neural networks (GNNs) encode molecules as graphs with atom and bond features, often extended to include 3D geometric information \citep{Gilmer2017, Schutt2017NIPS, Gasteiger2021}. Point-cloud models treat molecules as unordered sets of atomic coordinates, requiring networks that are invariant or equivariant to rotation and translation \citep{Fuchs2020, Thomas2018}.
Voxel-based 3D CNNs, which are directly relevant to this work, project atomic features onto a 3D grid and have been widely applied to pose prediction and binding-affinity scoring in docking pipelines \citep{Wallach2015, Ragoza2017, Jimenez2018KDEEP}. Unlike graph or point-based models that operate on discrete atomic representations, 3D CNNs can directly process continuous volumetric data such as electron density maps. Because our goal is to compare representational domains rather than model architectures, we adopt voxel-based CNNs as a consistent framework for learning from density fields. In contrast, GNNs and related architectures are designed for atom- and bond-level inputs and cannot directly handle volumetric electron-density data without new formulations that incorporate such information.

\paragraph{Electron Density in Structural Biology.}
Electron density maps are not merely alternative inputs for ML but are the fundamental data products of experimental techniques such as X-ray crystallography and cryo-EM, representing the time- and ensemble-averaged spatial distribution of electrons. Structural biologists interpret these maps to build and refine atomic models that best explain the observed density \citep{phenix}. This process involves fitting atomic templates, often requiring expert knowledge, and can be subjective—particularly in regions of lower resolution or higher flexibility. While essential for generating interpretable models, this step inherently involves assumptions and may lose subtle information present in the raw density. Using the density directly for ML bypasses this modeling step and can preserve more of the original experimental signal.

\paragraph{Machine Learning on Electron Density.}
Although less common than atom-based approaches, applying machine learning directly to electron density is an emerging area of research. Recent work has explored density maps for generative modeling—for example, \citet{Wang2022DensityGen} introduced a diffusion model that generates ligands conditioned on the electron density of protein pockets and showed that regions of rapid density change may correspond to non-covalent interactions. These findings motivate our use of both raw electron density and its gradient magnitude, which highlight complementary aspects of the electronic environment. However, few controlled studies have systematically compared density-based and atom-based representations across 3D molecular prediction tasks. Our work addresses this gap by directly evaluating both input types on two benchmark molecular prediction tasks.

\section{Experiments and Methods}
\label{sec:methods}

\subsection{Binding Affinity Prediction}
\label{sec:binding_affinity_pred}

We assess model performance using the PDBbind v2021 dataset \citep{pdbbind2021}, a widely used benchmark for protein-ligand binding affinity prediction, which includes $\sim$20,000 complexes with experimentally measured $pK$ values. Following \citet{pinheiro2024structure}, we voxelize the ligand and its surrounding protein pocket into separate 3D grids, both centered on the ligand’s center of mass, and pass them as input to the model (See Supplement \ref{supp:pdbbind_arch} for details).

To ensure proper generalization and avoid data leakage, we split the data using both receptor sequence and ligand similarity. Two complexes are assigned to the same split only if: (1) their receptor sequence identity exceeds 50\%, or (2) their receptor sequence identity exceeds 40\% and their ligand Tanimoto similarity is above 0.9. To further diversify the test set, all targets similar to those in the DEKOIS 2.0 benchmark \citep{dekois}—which spans a wide range of protein families—are reserved for the test set. The final split includes approximately 14,258 complexes for training, 1,171 for validation, and 5,554 for testing. 

We report Spearman correlation ($\rho$) on the test set as the primary evaluation metric. This choice reflects real-world applications of binding models, where compound ranking is often more critical than absolute value prediction.

\subsubsection{Input Feature Representations}

All inputs are voxelized into a $64 \times 64 \times 64$ grid at 0.25~\AA\ resolution. For density-based representations, we first resample all experimental electron density maps to a uniform resolution of 0.25~\AA, as raw maps vary in resolution across structures. We compare the following four input types:
\begin{itemize}
    \item \textbf{\texttt{Atom-Type}}: A multichannel representation where atom types are encoded as 3D Gaussians centered at atomic coordinates \citep{pinheiro2024structure}. Ligands use 7 channels (C, O, N, S, F, Cl, P), and protein pockets use 4 channels (C, O, N, S).
    \item \textbf{\texttt{Shape-Only}}: A single-channel baseline where all atoms are treated as carbon, removing chemical identity to focus on shape. 
    \item \textbf{\texttt{Density}}: A single-channel grid of experimental 2mFo–DFc electron density values, extracted from crystallographic MTZ files using Phenix \citep{phenix}.
    \item \textbf{\texttt{GradMag}}: A single-channel grid encoding the spatial gradient magnitude of the 2mFo–DFc map, highlighting regions of rapid density change, which may correlate with interaction sites \citep{Wang2022DensityGen}.
\end{itemize}

\subsubsection{Model Architectures and Training}

To isolate the effect of input representation, we use a consistent family of 3D convolutional networks across experiments. Models are evaluated at three capacity levels:
\begin{itemize}
    \item \textbf{\texttt{Tiny}} ($\sim$0.4M parameters): Based on GNINA’s \texttt{Default2018Affinity} architecture \citep{gnina}.
    \item \textbf{\texttt{Small}} ($\sim$4M) and \textbf{\texttt{Default}} ($\sim$58M): Use the encoder half of the VoxBind 3D U-Net architecture \citep{pinheiro2024structure}, followed by a multi-layer perceptron (MLP) for affinity prediction.

\end{itemize}
See Supplement~\ref{supp:pdbbind_arch} for detailed architectural descriptions.

All models are trained using the Adam optimizer (learning rate $1 \times 10^{-5}$), batch size 32, and mean squared error (MSE) loss, for 1000 epochs. Random 3D rotations are applied to voxel inputs during training to promote rotational invariance. Reported results are averaged over three random seeds.

\subsubsection{Data Efficiency Evaluation}

To assess data efficiency, we train models on increasingly larger subsets of the training set: 1\%, 5\%, 10\%, 25\%, 50\%, and 100\%. This allows us to compare performance trends across input representations and model sizes as data availability increases.

\subsection{Quantum Property Prediction}
We evaluate our models on the QM9 dataset \citep{qm9}, which contains approximately 134,000 small organic molecules (each with up to 9 heavy atoms), along with quantum chemical properties computed using Density Functional Theory (DFT). We train separate models to predict four scalar regression targets: dipole moment ($\mu$), isotropic polarizability ($\alpha$), energy of the highest occupied molecular orbital ($E_{\text{HOMO}}$), and energy of the lowest unoccupied molecular orbital ($E_{\text{LUMO}}$). The dataset is randomly split into 80\% training, 10\% validation, and 10\% test sets.

Unlike the PDBbind task, which leverages experimental electron density maps derived from crystallographic data, the QM9 dataset does not contain any experimental structural or density measurements. As a result, we generate approximate electron density maps using the GFN2-xTB semiempirical method via the XTB package \citep{grimme2017xtb}, based on the molecular geometries provided in the dataset. These computed densities are used as input for our density-based voxel representations.

We use mean absolute error (MAE) as the evaluation metric, following standard practice in prior work \citep{Schutt2017NIPS, Gasteiger2021}. Our goal is not to achieve state-of-the-art accuracy, but to evaluate how different representations affect predictive performance.

\subsubsection{Input Representations}

QM9 molecules are voxelized into $32 \times 32 \times 32$ grids at 0.25~\AA\ resolution, centered on the molecular center of mass. We use the same four input types as in the binding affinity task (Section~\ref{sec:binding_affinity_pred}): \texttt{Atom-Type}, \texttt{Shape-Only}, \texttt{Density}, and \texttt{GradMag}. The only difference lies in the atom-type representation: for QM9, we use five channels corresponding to C, H, O, N, and F. All other representations use a single channel.

\subsubsection{Model Architectures and Training}

To study the effect of model capacity, we use three versions of a 3D CNN architecture adapted from VoxMol \citep{pinheiro20243dmoleculegenerationdenoising}, varying in size: \texttt{Tiny} ($\sim$4M parameters), \texttt{Small} ($\sim$15M), and \texttt{Default} ($\sim$58M). For architectural details, see Supplement~\ref{supp:qm9_arch}.

All models are trained using the Adam optimizer with a learning rate of $1 \times 10^{-5}$, batch size 128, and Mean Squared Error (MSE) loss. To improve invariance to molecular orientation, we apply random 3D rotations to voxelized inputs during training.

Prior to training, all target labels are normalized to zero mean and unit variance. Predictions are rescaled during evaluation to report metrics in the original units. Each model is initially trained for 1500 epochs. Training continues for up to 5000 epochs if the validation loss has not converged, using early stopping with a patience of 50 epochs. Final results are reported as the mean and standard deviation of the MAE across three runs with different random seeds.

\subsubsection{Data Efficiency Evaluation}

To evaluate data efficiency, we train the models on subsets of the training data: 0.15\%, 1\%, 10\%, and 100\%. Performance across these subsets helps assess how each input representation scales with data availability.

\section{Results}
\label{sec:results}

\subsection{Binding Affinity Prediction (PDBbind)}

\begin{figure*}[t]
\vskip 0.2in
\centering
\centerline{\includegraphics[width=\textwidth]{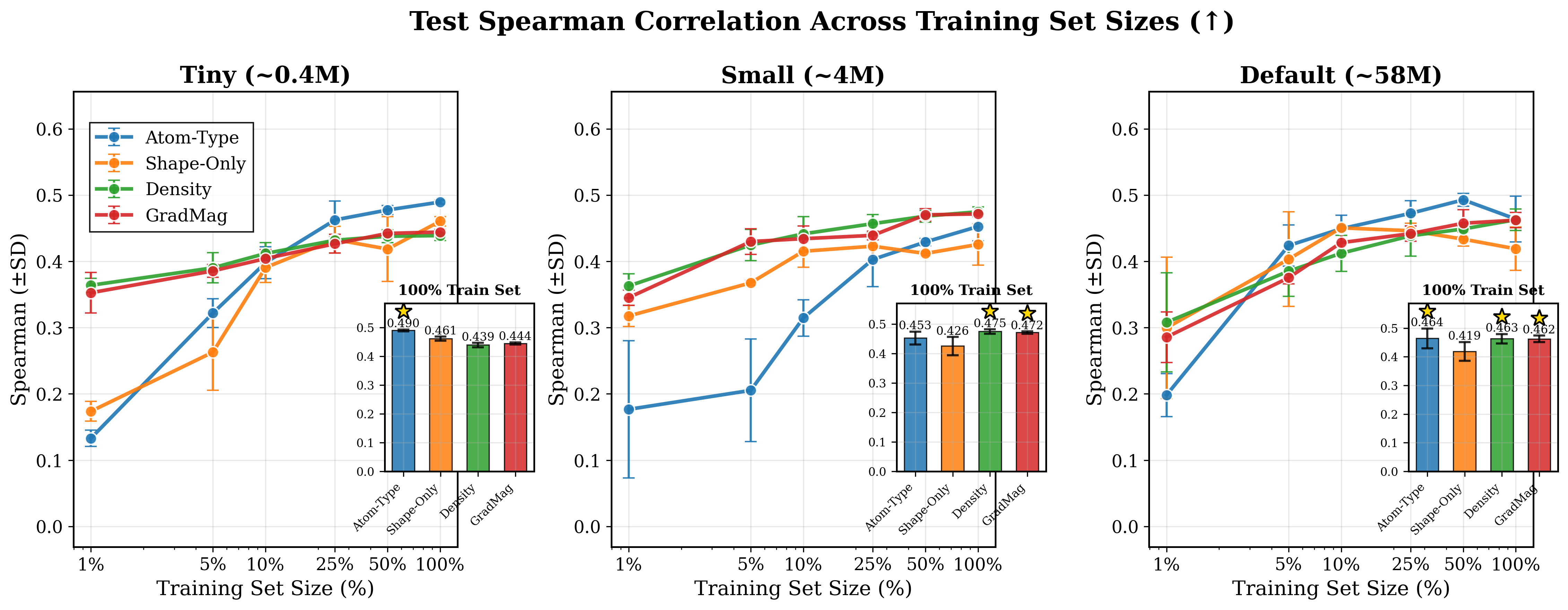}}
\caption{Test Spearman correlation for binding affinity prediction on the PDBbind test set across training set sizes (1\%–100\%) and model capacities ($\sim$0.4M, $\sim$4M, $\sim$58M parameters).
We compare four voxel-based input representations: atom types (\texttt{Atom-Type}), atoms mapped to carbon (\texttt{Shape-Only}), electron density values (\texttt{Density}), and electron density gradient magnitude (\texttt{GradMag}). Performance generally improves with training data size but plateaus beyond 10\%. With the full training set, all representations perform similarly across model sizes. In the low-data regime (1\%), density-based inputs outperform \texttt{Atom-Type}, and the \texttt{Shape-Only} baseline—despite discarding chemical identity—performs comparably to density-based inputs. This counterintuitive result suggests that simple spatial occupancy alone may be highly predictive in this dataset, potentially due to biases in the benchmark or the use of static, bound structures. Insets show performance at 100\% training data, with stars marking best-performing models within standard deviation.}
\label{fig:binding_affinity_model_size_spearman}
\vskip -0.2in
\end{figure*}

Figure~\ref{fig:binding_affinity_model_size_spearman} shows the Spearman correlation on the PDBbind test set across training set sizes and model capacities. We observe that increasing model size has little effect on performance. The \texttt{Tiny} model ($\sim$0.4M parameters) performs similarly to the largest model (\texttt{Default}, $\sim$58M), consistent with previous findings that small architectures can perform well in structure-based docking and virtual screening tasks \citep{gnina}.

Model performance improves with more training data but plateaus after 10\%. At full data scale, all four input representations—\texttt{Atom-Type}, \texttt{Shape-Only}, \texttt{Density}, and \texttt{GradMag}—perform similarly within error margins.

Interestingly, in the low-data regime (1\% of the training set, or $\sim$100 complexes), the density-based inputs (\texttt{Density} and \texttt{GradMag}) already achieved relatively strong performance---close to their performance when trained on the full dataset---and outperformed the atom-type input (\texttt{Atom-Type}) across all models. Even more surprisingly, the \texttt{Shape-Only} input, which treats all atoms as carbon and discards atomic identity, performed comparably to the density-based inputs at this data size (except for the \texttt{Tiny} model). This result is counter-intuitive: one might expect \texttt{Shape-Only} to perform the worst as we removing the atom types. However, these findings suggest that in this dataset---where we use experimentally resolved structures in their bound, low-energy conformations---spatial occupancy alone (i.e., how well the ligand fills the binding pocket) may be a strong predictive signal. Prior work has noted that hydrophobic and shape-complementary interactions are often the dominant contributors to binding affinity, while atom-type–specific interactions (e.g., hydrogen bonds, salt bridges) tend to govern binding specificity \citep{bissantz2010medicinal}.

The small difference between 10\% and 100\% training data further supports this: models may already extract most of the relevant information early on. In contrast, the \texttt{Atom-Type} input performs worse in low-data settings, likely due to its higher dimensionality (7 channels for ligand atoms and 4 for protein atoms), leading to sparser inputs and greater risk of overfitting.

\subsection{Quantum Property Prediction}

\begin{figure*}[t]
\vskip 0.2in
\centering
\centerline{\includegraphics[width=\textwidth]{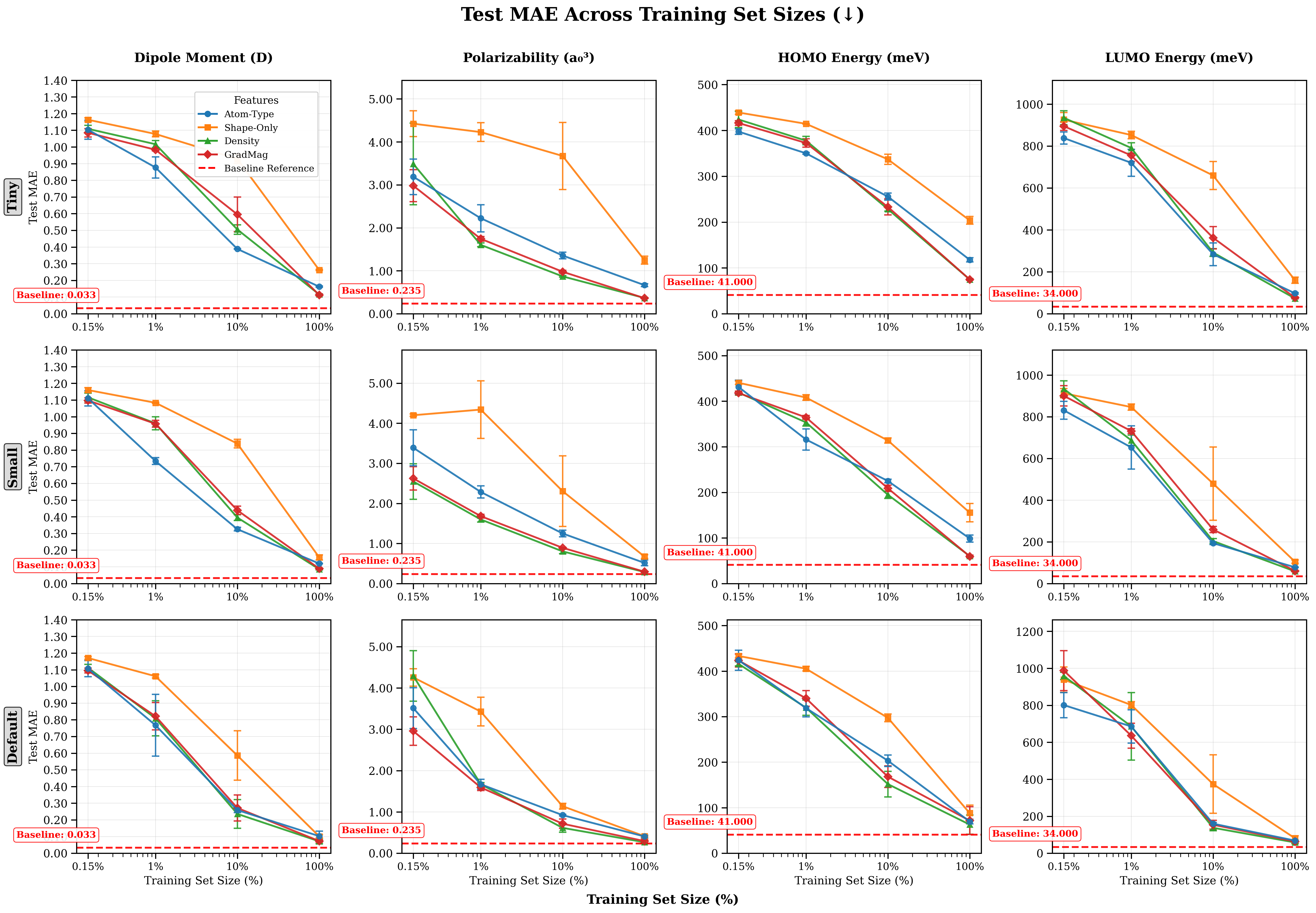}}
\caption{Test MAE across training set sizes for QM9 target properties. Each row corresponds to a different model size (\texttt{Tiny} $\sim$4M, \texttt{Small} $\sim$15M, \texttt{Default} $\sim$58M), and each column to a regression target (Dipole Moment, Polarizability, HOMO Energy, LUMO Energy). Lower MAE indicates better performance. Across all settings, increasing training data consistently reduces error. Density-based inputs (\texttt{Density}, \texttt{GradMag}) outperform atom-based ones at full data scale, while the poor performance of \texttt{Shape-Only} (orange) highlights the value of chemically informative features. Red dashed lines mark reported SchNet \citep{Schutt2017NIPS} results, shown only as a sanity check to confirm that our 3D CNN benchmarks yield reasonable error ranges. Our models were not tuned for state-of-the-art performance—the goal is to compare voxel-based representations under consistent conditions. Because SchNet is a graph neural network operating on atom-level graphs, it cannot directly represent volumetric density data without substantial reformulation, and its results are therefore not directly comparable to ours.}

\label{fig:qm9_test_mae_best_ckpt}
\vskip -0.2in
\end{figure*}

Figure~\ref{fig:qm9_test_mae_best_ckpt} shows test MAE across training set sizes for QM9 target properties. As expected, performance improves consistently with more training data, regardless of model size, input type, or prediction target. This contrasts with the PDBbind results, where performance plateaus after 10\%, and highlights QM9's greater sensitivity to data quantity.

\begin{figure*}[t]
\vskip 0.2in
\centering
\centerline{\includegraphics[width=\textwidth]{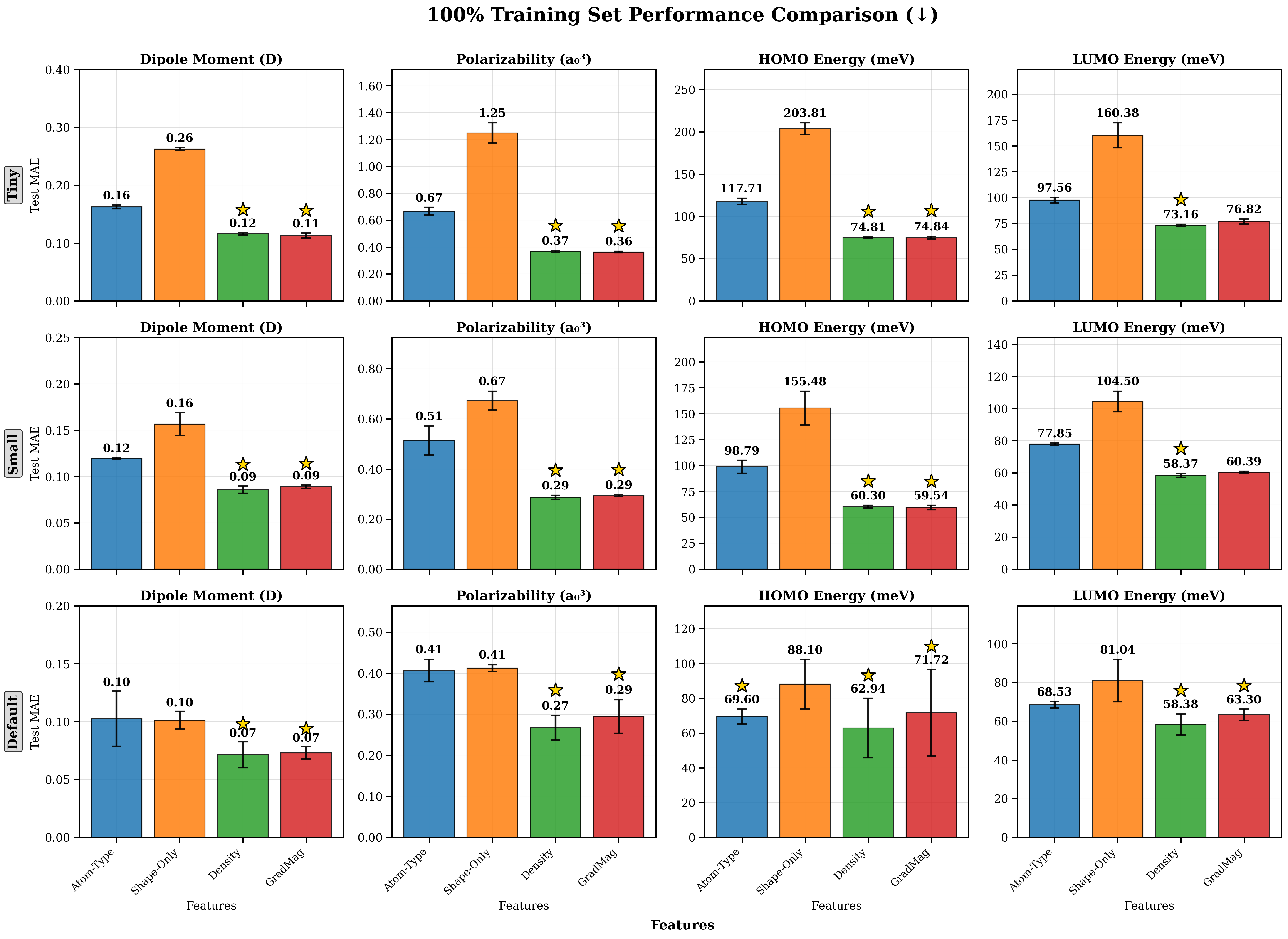}}
\caption{Test MAE at 100\% training set size for all input types, targets, and model sizes. Each row shows results for a different model size, and each column corresponds to one of the QM9 target properties. Bars show mean test MAE with standard deviation across three seeds. Density-based inputs (\texttt{Density}, \texttt{GradMag}) consistently yield the lowest errors across all targets and model sizes. \texttt{Shape-Only} inputs perform worst overall, highlighting the value of chemically meaningful information in voxel inputs.
}
\label{fig:qm9_test_mae_barplot_100}
\vskip -0.2in
\end{figure*}

Figure~\ref{fig:qm9_test_mae_barplot_100} summarizes results at 100\% training data. Accuracy improves steadily with model size—from \texttt{Tiny} ($\sim$4M) to \texttt{Small} ($\sim$15M) and \texttt{Default} ($\sim$58M)—unlike the binding task, where model size had little effect. Atom-type information plays a more important role here: removing atomic identity (\texttt{Shape-Only}) consistently reduces performance across all models and data sizes, in line with the expectation that quantum properties are driven by electronic structure, not just shape.

In low-data regimes, performance varies across input types with no clear winner. However, at full data scale, density-based inputs (\texttt{Density}, \texttt{GradMag}) consistently outperform atom-type representations. These densities are generated using the semiempirical XTB method, which is significantly less computationally expensive—but also less accurate—than the DFT calculations used to derive the target molecular properties. This mismatch in the level of theory introduces a potential source of error; for example, XTB tends to systematically overestimate dipole moments relative to DFT (see Supplement, Figure~\ref{fig:dipole-error-distribution}).

Additional approximation error comes from voxelizing continuous densities. The supplemental figure reports error based on continuous basis-function densities, whereas our models use discretized voxel inputs, which add further discrepancy. Despite these limitations, \texttt{Density} and \texttt{GradMag} outperform both \texttt{Atom-Type} and \texttt{Shape-Only} (Figure~\ref{fig:qm9_test_mae_barplot_100}), suggesting that voxelized density still captures important aspects of electronic structure not present in other representations.

\paragraph{Note on external comparisons.} We do not compare to prior PDBbind models, as published results use different data splits and are not directly comparable without retraining. Moreover, recent work has shown that the splits used in previous studies may suffer from data leakage due to high sequence similarity between training and test proteins \citep{li2024leak, graber2024gems}. Our experiments use controlled splits with minimal train-test overlap to fairly assess the effect of input representations (see Section \ref{sec:binding_affinity_pred}). For QM9, we include SchNet \citep{Schutt2017NIPS} performance (Figure~\ref{fig:qm9_test_mae_best_ckpt}) only as a sanity check to verify that our 3D CNN benchmarks produce results within a reasonable error range. Our CNN models were not further tuned or optimized for performance, as the goal is to compare voxel-based representations under consistent training conditions, rather than to outperform graph-based or other state-of-the-art methods published in the literature. Because SchNet operates on graph-based atomic representations rather than volumetric grids, its results are not directly comparable to our voxel-based models.

\section{Conclusion}
\label{sec:conclusion}

This study systematically evaluated voxel-based molecular representations—\texttt{Atom-Type}, electron density (\texttt{Density}), and density gradient magnitude (\texttt{GradMag})—as inputs to 3D convolutional neural networks for molecular property prediction. Motivated by the hypothesis that density-derived representations offer a richer, more physically grounded encoding than discrete atom-type models, we benchmarked these approaches across two prediction tasks: 3D binding affinity prediction using experimental protein–ligand structures (PDBbind) \citep{pdbbind2021}, and quantum property prediction of small molecules (QM9) \citep{qm9}.

For 3D binding affinity prediction, increasing model size—from $\sim$0.4M to $\sim$58M parameters—had minimal impact on performance, consistent with prior work showing the effectiveness of compact architectures (e.g., GNINA \citep{gnina}) for structure-based drug discovery. Model accuracy improved with additional training data but plateaued beyond $\sim$10\%, indicating that most predictive information is captured early. In the low-data regime (1\%), density-based inputs (\texttt{Density}, \texttt{GradMag}) outperformed atom-type inputs and achieved near-peak performance. Interestingly, a simplified input that removes atom-type information (\texttt{Shape-Only}) performed comparably to density-based inputs. These findings suggest that in this setup—using bound-state experimental structures where ligand poses and steric complementarity are already resolved—geometric occupancy alone may provide a strong predictive signal. In such cases, atom-type information may add limited benefit and can increase overfitting risk in low-data settings due to higher input sparsity.

In contrast, the quantum property prediction task on QM9 exhibited markedly different behavior. Performance improved consistently with both model capacity and data scale, with larger networks yielding better accuracy, indicating a greater need for representational expressiveness in this setting. Atom-type information also played a more critical role: removing atom identity (\texttt{Shape-Only}) substantially degraded performance across all targets and model sizes, reflecting the importance of chemical specificity in quantum behavior. While no representation was consistently superior in low-data settings, density-based inputs (\texttt{Density}, \texttt{GradMag}) consistently outperformed atom-type inputs when trained on the full dataset. These densities were computed using the semiempirical XTB method, which is significantly less computationally expensive—but also less accurate—than the DFT methods used to generate the QM9 target properties. This introduces two distinct sources of approximation: (1) differences in the level of theory—e.g., XTB systematically overestimates dipole moments relative to DFT (see Supplement, Figure~\ref{fig:dipole-error-distribution})—and (2) discretization error from voxelizing continuous densities. Despite these limitations, density-based inputs yielded the best performance overall, suggesting that electron density captures essential information about electronic structure that is not easily recovered from atom types or spatial geometry alone.

While these findings highlight the strengths of density-based voxel representations, several limitations and opportunities remain. This study focuses on 3D CNNs because they are the most natural choice for processing volumetric electron density data from both experimental and computed sources. Extending these ideas to graph-based or equivariant architectures would require new formulations capable of representing density within their atom-centric frameworks. Moreover, voxel-based representations are computationally intensive—storing and training on high-resolution volumetric grids can be prohibitively expensive for large datasets. Developing more compact or adaptive encodings could make density-based learning more practical at scale. It would also be valuable to explore hybrid approaches that combine atom-type specificity with density-derived features, potentially capturing complementary geometric and electronic information. Ultimately, the optimal molecular representation depends on the prediction task, data regime, and underlying physical principles most relevant to the property being modeled.

\bibliographystyle{unsrtnat}
\bibliography{bibliography}


\newpage

\appendix

\section{Model Architectures}

\subsection{Binding Affinity Model Architecture (PDBbind)}
\label{supp:pdbbind_arch}

We evaluate three 3D CNN architectures for predicting protein–ligand binding affinity using the PDBbind v2021 dataset \citep{pdbbind2021}. Each complex is voxelized into two separate $64^3$ grids—one for the ligand and one for the protein pocket—centered on the ligand’s center of mass and resampled to 0.25~\AA\ resolution.

\paragraph{Input Representations.}
Channel dimensions vary based on the representation:
\begin{itemize}
    \item \texttt{Atom-Type}: 7 channels for ligand atoms (C, O, N, S, F, Cl, P), 4 channels for protein pocket atoms (C, O, N, S).
    \item \texttt{Shape-Only}, \texttt{Density}, \texttt{GradMag}: 1 channel each for ligand and protein pocket atoms.
\end{itemize}

\paragraph{Ligand and Pocket Encoders.}
The ligand and pocket are processed independently using identical 3D CNN encoders adapted from VoxBind \citep{pinheiro2024structure}. Each encoder consists of a single residual block comprising two padded $3 \times 3 \times 3$ convolutional layers with 16 channels, followed by \texttt{SiLU} activations \citep{elfwing2018silu}. Each produces an output of shape $16 \times 64^3$. The two outputs are summed element-wise to produce a fused embedding.

\paragraph{Fused Representation Encoder.}
The fused representation is then processed by one of the following architectures depending on model capacity:
\begin{itemize}
    \item \textbf{\texttt{Tiny}} ($\sim$0.4M parameters): The fused embedding is passed through the GNINA \texttt{Default2018} CNN architecture \citep{gnina}. This consists of five 3D convolutional layers with interleaved average pooling and \texttt{ReLU} activations \citep{relu}, followed by a fully connected linear layer to produce a scalar affinity prediction.

    \item \textbf{\texttt{Small}} ($\sim$4M parameters): The fused embedding is passed through the encoder portion of the 3D U-Net from VoxBind \citep{pinheiro2024structure}. This U-Net encoder follows the original VoxBind design, using four resolution levels with channel multipliers \texttt{[1, 2, 2, 4]} and base channel width $n_\text{ch} = 8$. Each resolution level contains two residual blocks, and each residual block consists of two padded $3 \times 3 \times 3$ convolutions with 16 channels followed by \texttt{SiLU} activations \citep{elfwing2018silu}. Group normalization \citep{wu2018group} with 4 groups is applied throughout. The encoder outputs a bottleneck feature map of shape $128 \times 8^3$.

    \item \textbf{\texttt{Default}} ($\sim$58M parameters): Same as \texttt{Small}, but with $n_\text{ch} = 32$, yielding a bottleneck of $512 \times 8^3$. Group normalization uses 16 groups.
\end{itemize}

\paragraph{MLP Prediction Head (Small and Default only).}
The bottleneck feature map is passed through a shared multi-layer perceptron (MLP) head to produce the final affinity prediction. This consists of:
\begin{itemize}
    \item \texttt{AvgPool3d}(8)
    \item Linear layers with \texttt{LayerNorm}, \texttt{SiLU}, and \texttt{Tanhshrink} activations:
    \[
    \texttt{Small: } 128 \rightarrow 128 \rightarrow 64 \rightarrow 1
    \]
    \[
    \texttt{Default: } 512 \rightarrow 512 \rightarrow 64 \rightarrow 1
    \]
\end{itemize}

\subsection{Quantum Property Prediction Model Architecture (QM9)}
\label{supp:qm9_arch}

We use a 3D CNN encoder adapted from VoxMol \citep{pinheiro20243dmoleculegenerationdenoising} for scalar regression on voxelized QM9 molecules \citep{qm9}. Each molecule is represented as a $32^3$ grid.

\paragraph{Input Representations.}
\begin{itemize}
    \item \texttt{Atom-Type}: 5 channels (H, C, N, O, F)
    \item \texttt{Shape-Only}, \texttt{Density}, \texttt{GradMag}: 1 channel each
\end{itemize}

\paragraph{Encoder.}
The input is first projected to $n_\text{ch}$ base channels using a $3 \times 3 \times 3$ convolution. The encoder applies four downsampling stages using channel multipliers \texttt{[1, 2, 2, 4]}, resulting in a final output of size $16n_\text{ch} \times 4^3$. Each stage contains two residual blocks with 3D convolution, \texttt{GroupNorm}, and \texttt{SiLU} activations. Self-attention is used in the final two stages.

\paragraph{MLP Prediction Head.}
The encoder output is passed to a multi-layer perceptron for scalar quantum property prediction:
\[
16n_\text{ch} \rightarrow 64 \rightarrow 32 \rightarrow 1
\]
Each layer includes \texttt{LayerNorm}, \texttt{SiLU}, and \texttt{Tanhshrink} activations.

\paragraph{Model Variants.}
The QM9 models share this architecture but vary in channel width and normalization:
\begin{itemize}
    \item \textbf{\texttt{Tiny}} ($\sim$4M): $n_\text{ch} = 8$, $n_\text{groups} = 4$
    \item \textbf{\texttt{Small}} ($\sim$15M): $n_\text{ch} = 16$, $n_\text{groups} = 8$
    \item \textbf{\texttt{Default}} ($\sim$58M): $n_\text{ch} = 32$, $n_\text{groups} = 16$
\end{itemize}

\section{Distribution of Dipole Moment Errors: XTB vs DFT}
\label{supp:dipole_error}

\begin{figure}[h]
    \centering
    \includegraphics[width=0.75\linewidth]{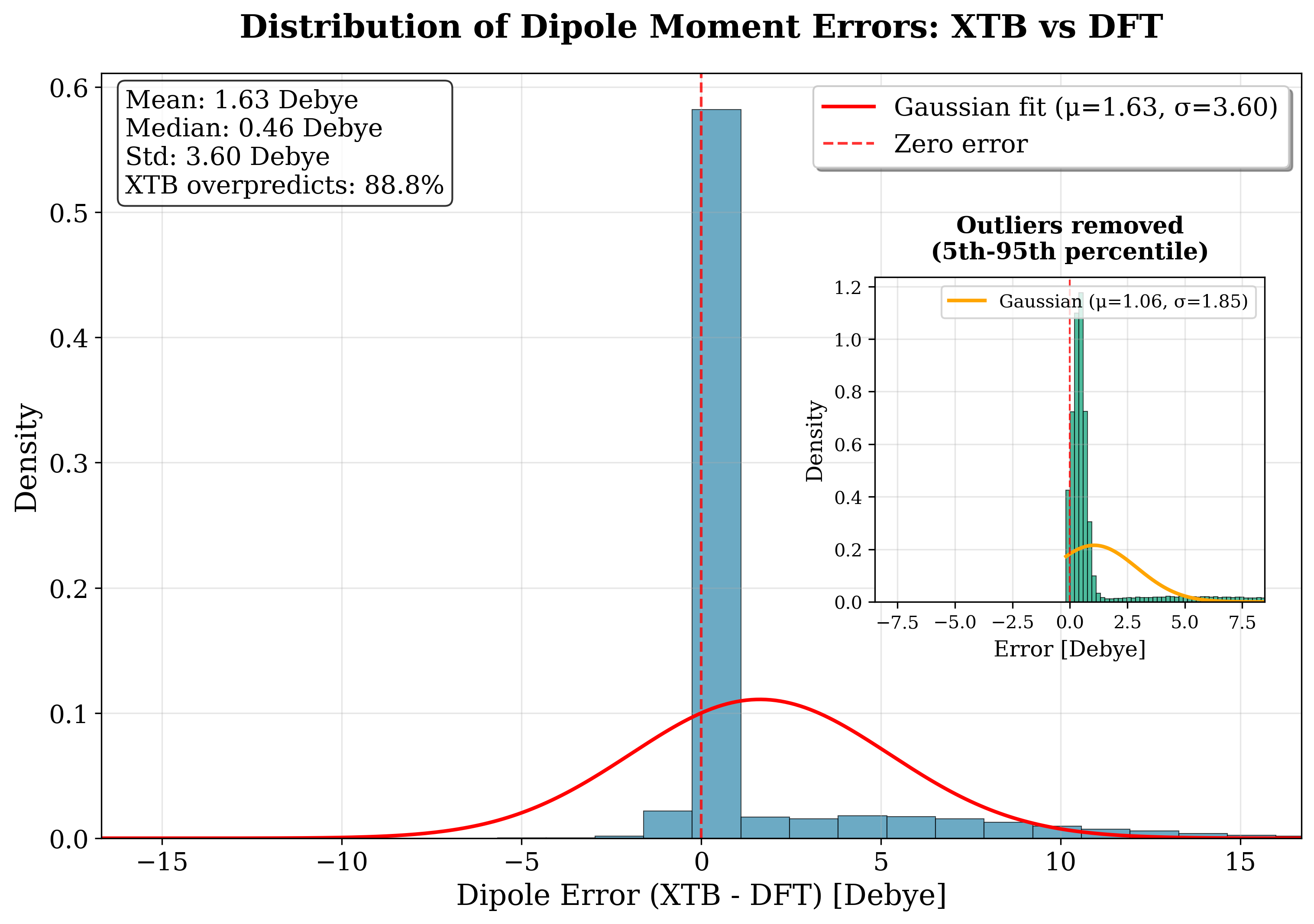}
    \caption{\textbf{Distribution of Dipole Moment Errors: XTB vs DFT.}
    Histogram of dipole moment errors (XTB $-$ DFT) across the QM9 dataset. The red solid line shows a Gaussian fit to the full error distribution ($\mu = 1.63$, $\sigma = 3.60$ Debye), while the dashed red line indicates zero error. The inset shows the distribution with outliers removed (5th–95th percentile), highlighting the skew and overprediction tendency of XTB. XTB overpredicts dipole moments in 88.8\% of cases.}
    \label{fig:dipole-error-distribution}
\end{figure}








\end{document}